\documentclass[conference]{IEEEtran}
\IEEEoverridecommandlockouts
\usepackage{cite}
\usepackage{amsmath,amssymb,amsfonts}
\usepackage{algorithmic}
\usepackage{graphicx}
\usepackage{textcomp}
\usepackage{xcolor}
\usepackage{multirow}
\usepackage{fancyhdr}
\def\BibTeX{{\rm B\kern-.05em{\sc i\kern-.025em b}\kern-.08em
    T\kern-.1667em\lower.7ex\hbox{E}\kern-.125emX}}

\fancypagestyle{firstpage}{
  \fancyhf{}
  
  \fancyfoot[L]{\footnotesize © 2023 IEEE. Personal use of this material is permitted. Permission from IEEE must be obtained for all other uses, in any current or future media, including reprinting/republishing this material for advertising or promotional purposes, creating new collective works, for resale or redistribution to servers or lists, or reuse of any copyrighted component of this work in other works. DOI: 10.1109/ICAICTA59291.2023.10390334}
}

\begin{document}

\title{Supervised Learning Models for Early Detection of Albuminuria Risk in Type-2 Diabetes Mellitus Patients}

\author{\IEEEauthorblockN{1\textsuperscript{st} Arief Purnama Muharram\textsuperscript{*}}
\IEEEauthorblockA{\textit{Master Program of Informatics} \\
\textit{School of Electrical Engineering and Informatics} \\
\textit{Institut Teknologi Bandung} \\
Bandung, Indonesia \\
23521013@std.stei.itb.ac.id \\ \thanks{\textsuperscript{*}Corresponding author}}
\and
\IEEEauthorblockN{2\textsuperscript{nd} Dicky Levenus Tahapary\textsuperscript{*}}
\IEEEauthorblockA{\textit{Metabolic Disorder, Cardiovascular and Aging Research Center} \\
\textit{Indonesia Medical Education and Research Institute} \\
\textit{Faculty of Medicine Universitas Indonesia} \\
Jakarta, Indonesia \\
dicky.tahapary@ui.ac.id}
\and
\IEEEauthorblockN{3\textsuperscript{rd} Yeni Dwi Lestari}
\IEEEauthorblockA{\textit{Ophthalmology Department} \\
\textit{Faculty of Medicine Universitas Indonesia} \\
\textit{Cipto Mangunkusumo General Hospital}\\
Jakarta, Indonesia \\
yeni.lestari@ui.ac.id}
\and
\IEEEauthorblockN{4\textsuperscript{th} Randy Sarayar}
\IEEEauthorblockA{\textit{Faculty of Medicine} \\
\textit{Universitas Indonesia}\\
Jakarta, Indonesia \\
randy.sarayar91@ui.ac.id}
\and
\IEEEauthorblockN{5\textsuperscript{th} Valerie Josephine Dirjayanto}
\IEEEauthorblockA{\textit{Faculty of Medicine} \\
\textit{Universitas Indonesia}\\
Jakarta, Indonesia \\
valerie.josephine@ui.ac.id}
}

\maketitle

\thispagestyle{firstpage}

\begin{abstract}
Diabetes, especially T2DM, continues to be a significant health problem. One of the major concerns associated with diabetes is the development of its complications. Diabetic nephropathy, one of the chronic complication of diabetes, adversely affects the kidneys, leading to kidney damage. Diagnosing diabetic nephropathy involves considering various criteria, one of which is the presence of a pathologically significant quantity of albumin in urine, known as albuminuria. Thus, early prediction of albuminuria in diabetic patients holds the potential for timely preventive measures. This study aimed to develop a supervised learning model to predict the risk of developing albuminuria in T2DM patients. The selected supervised learning algorithms included Naïve Bayes, Support Vector Machine (SVM), decision tree, random forest, AdaBoost, XGBoost, and Multi-Layer Perceptron (MLP). Our private dataset, comprising 184 entries of diabetes complications risk factors, was used to train the algorithms. It consisted of 10 attributes as features and 1 attribute as the target (albuminuria). Upon conducting the experiments, the MLP demonstrated superior performance compared to the other algorithms. It achieved accuracy and f1-score values as high as 0.74 and 0.75, respectively, making it suitable for screening purposes in predicting albuminuria in T2DM. Nonetheless, further studies are warranted to enhance the model's performance.
\end{abstract}


\section{Introduction}
Diabetes continues to be one of the most challenging noncommunicable diseases worldwide. It is a chronic metabolic disorder characterized by high blood sugar levels caused by problems in insulin production, sensitivity of cells' response to insulin, or both \cite{perkeni2021}. There are four types of diabetes, namely type-1, type-2, gestational type, and other types. However, type-2 diabetes (T2DM) dominates all other diabetes types \cite{idf2021}, accounting for more than 90\% of all diabetes cases. The high prevalence of T2DM is strongly associated with the unhealthy modern lifestyle, including unhealthy eating habits, smoking, obesity, and a lack of physical activity, as well as internal predisposition factors such as race and family history \cite{kyoru2020}.

The predominant challenge associated with diabetes stems from the array of complications that can arise when diabetes is not adequately controlled. Among these unwanted complications, one particularly notable issue is kidney complication, which falls under the category of microvascular complications, affecting the smaller blood vessels \cite{chamine2022,haas1993}. This specific complication is commonly referred to as diabetic nephropathy and accounts for approximately 14.0\% of diabetes-related complications \cite{chamine2022}. 

Diabetic nephropathy is considered as a type of Chronic Kidney Disease (CKD). According to the Kidney Disease Improving Global Outcomes (KDIGO) 2012 guidelines, CKD is established when there are markers of kidney damage and/or a Glomerular Filtration Rate (GFR) $<$ 60 mL/min/1.73m\textsuperscript{2} that lasts for at least \text{$\ge$} 3 months. The kidney damage markers for CKD include the presence of pathologically high quantities of urinary albumin excretion (albuminuria), the presence of urine sediment abnormalities, structural abnormalities detected by imaging, and a history of kidney transplantation \cite{kdigo2012}.

As mentioned in the preceding paragraph, the presence of albuminuria can be indicative of a kidney problem. Albumin in urine can signal an issue with the kidney filtration function. Albuminuria can be divided into two categories: microalbuminuria and macroalbuminuria. Microalbuminuria is diagnosed when the albumin-creatinine ratio is $>$ 30 mg/24h and $<$ 300 mg/24h, while macroalbuminuria is diagnosed when the albumin excretion is $>$ 300 mg/24h in a 24-hour urine collection sample \cite{pavkov2018}. As albuminuria can serve as a signal of kidney problems, it becomes essential for diabetes patients to be aware of their risk of developing this condition.

Therefore, the primary objective of this study is to develop a supervised learning model capable of predicting the risk of albuminuria development in diabetes patients, particularly those with T2DM. The primary contributions of this paper can be summarized as follows:
\begin{itemize}
    \item Development of a supervised model capable of predicting early albuminuria in patients with type 2 diabetes mellitus (T2DM).
    \item Identification of the optimal supervised algorithm for early albuminuria detection in T2DM patients.
\end{itemize}

\section{Related Work}
Recently, there has been a growing interest among researchers in using machine learning approaches to predict albuminuria. This interest arise from the urgency of developing early risk prediction tools for the disease, as it can lead to increased "costs" if left undetected. To our knowledge, two studies conducted by Khitan et al. \cite{khitan2021} and Lin et al. \cite{lin2022} have used machine learning approaches for predicting albuminuria.

Khitan et al. \cite{khitan2021} in their study used machine learning approaches to predict the risk of albuminuria in person with diabetes. Their study incorporated 13 predictive factors, including measures such as subtotal lean mass, subtotal fat mass, diabetes duration, age, HbA1c levels, creatinine levels, triglyceride levels, total cholesterol levels, HDL cholesterol levels, maximum exercise capacity, systolic and diastolic blood pressure, and ankle brachial index. They conducted their study on 1330 subjects and used a variety of machine learning algorithms, including random forest, gradient boost, logistic regression, support vector machines, multilayer perceptron, and a stacking classifier. The results showed that the multilayer perceptron (MLP) exhibited the highest performance with an AUC (Area Under the Curve) value of 0.67. Furthermore, the model demonstrated a precision of 0.61, recall of 0.67, and an accuracy of 0.62, as determined from the confusion matrix presented in the paper.

In another study within this domain, Lin et al. \cite{lin2022} aimed to predict microalbuminuria in the Chinese population using machine learning approaches. Their study involved 3,294 subjects ranging in age from 16 to 93 years. They used the "glm" package in the R software to construct their machine learning model. Their model achieved a specificity of 0.9 and an accuracy of 0.63, although the sensitivity was relatively low at 0.2. Despite these outcomes, the study's conclusions highlighted systolic and diastolic blood pressure, fasting blood glucose levels, triglyceride levels, gender, age, and smoking as potential predictors of microalbuminuria among the patient population.

\section{Methodology}
\subsection{Dataset}

\begin{figure}[htbp]
\centerline{\includegraphics[width=0.5\textwidth]{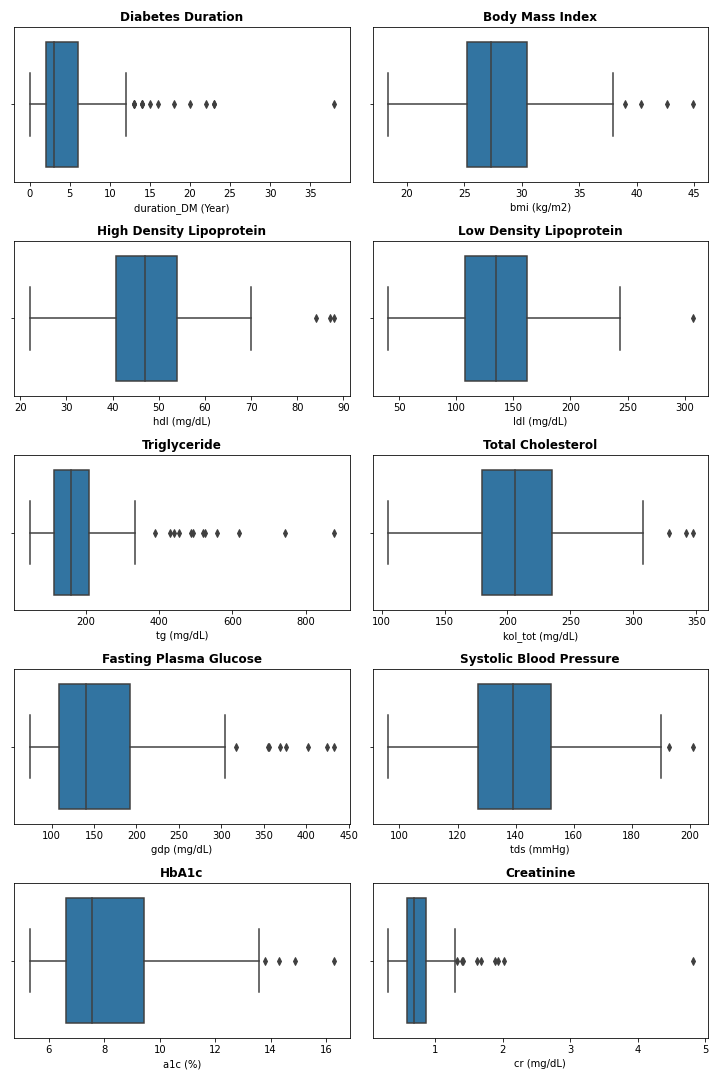}}
\caption{Dataset distribution}
\label{fig:dataset_boxplot}
\end{figure}

\begin{table}[htbp]
\caption{Summary Statistics}
\begin{center}
\begin{tabular}{|c|c|c|c|c|c|}
\hline
\textbf{Attribute} & \textbf{Unit} & \textbf{Mean$\pm$Std} & \textbf{Min} & \textbf{Max} \\ \hline
\text{durasi\_dm} & \text{Year} & \text{5.168$\pm$5.012} & \text{0.000} & \text{38.000} \\ \hline
\text{bmi} & \text{kg/m\textsuperscript{2}} & \text{27.973$\pm$4.517} & \text{18.350} & \text{44.870} \\ \hline
\text{hdl} & \text{mg/dL} & \text{48.065$\pm$10.913} & \text{22.000} & \text{88.000} \\ \hline
\text{ldl} & \text{mg/dL} & \text{135.978$\pm$38.522} & \text{40.000} & \text{307.000} \\ \hline
\text{tg} & \text{mg/dL} & \text{185.571$\pm$118.305} & \text{47.000} & \text{878.000} \\ \hline
\text{kol\_tot} & \text{mg/dL} & \text{207.554$\pm$42.534} & \text{105.000} & \text{347.000} \\ \hline
\text{gdp} & \text{mg/dL} & \text{158.076$\pm$68.591} & \text{74.000} & \text{433.000} \\ \hline
\text{TDS} & \text{mmHg} & \text{141.609$\pm$18.682} & \text{96.000} & \text{201.000} \\ \hline
\text{a1c} & \text{\%} & \text{8.204$\pm$2.134} & \text{5.300} & \text{16.300} \\ \hline
\text{cr} & \text{mg/dL} & \text{0.782$\pm$0.409} & \text{0.300} & \text{4.810} \\ \hline
\textbf{Attribute} & \multicolumn{2}{|c|}{\textbf{Category}} & \textbf{Count} & \textbf{Perc.} \\ \hline
\text{kid\_group} & \multicolumn{2}{|c|}{Normal (0)} & \text{92} & \text{0.50} \\
& \multicolumn{2}{|c|}{Albuminuria (1)} & \text{92} & \text{0.50} \\ \hline
\end{tabular}
\label{tab:dataset_summary_statistics}
\end{center}
\end{table}

In this study, we used our private dataset consisting of data on the risk of diabetes complications, which was collected from a primary healthcare facility in DKI Jakarta, Indonesia. The dataset comprises 184 records, each consisted of 10 features and 1 target variable (Table \ref{tab:dataset_summary_statistics}) (Figure \ref{fig:dataset_boxplot}). All records are sourced from patients with T2DM. The features are all numerical, whereas the target variable is categorical. Prior to analysis, all data were carefully examined and cleaned to remove any missing values or measurement errors. To ensure the security and privacy of the medical data, all information was anonymized.
\begin{itemize}
    \item \textbf{durasi\_dm:} This attribute refers to the length of time since the patient's initial diabetes diagnosis. The duration is measured in years.
    \item \textbf{bmi:} This attribute refers to the patient's current Body Mass Index (BMI), which is measured in kg/m\textsuperscript{2}.
    \item \textbf{hdl:} This attribute refers to the current level of High-Density Lipoprotein (HDL) in the bloodstream, measured in mg/dL using standard laboratory methods.
    \item \textbf{ldl:} This attribute refers to the current level of Low-Density Lipoprotein (LDL) in the bloodstream, measured in mg/dL using standard laboratory methods.
    \item \textbf{tg:} This attribute refers to the current level of triglyceride in the bloodstream, measured in mg/dL using standard laboratory methods.
    \item \textbf{kol\_tot:} This attribute refers to the current level of total cholesterol in the bloodstream, measured in mg/dL using standard laboratory methods.
    \item \textbf{gdp:} This attribute refers to the current level of fasting plasma glucose in the bloodstream, measured in mg/dL using standard laboratory methods.
    \item \textbf{TDS:} This attribute refers to the current systolic blood pressure measured in mmHg using an ambulatory blood pressure device.
    \item \textbf{a1c:} This attribute refers to the current level of HbA1c measured using standard laboratory methods.
    \item \textbf{cr:} This attribute refers to the current level of creatinine, measured in mg/dL using standard laboratory methods.
    \item \textbf{kid\_group:} This attribute serves as the target label and describes the grouping of kidney disease. It is a categorical attribute comprising of two categories: normal and albuminuria. The determination of albuminuria label was based on the KDIGO 2012 criteria. However, instead of treating microalbuminuria and macroalbuminuria as separate categories, we classified them both under the umbrella term of "albuminuria".
\end{itemize}

\begin{figure}[htbp]
\centerline{\includegraphics[width=0.5\textwidth]{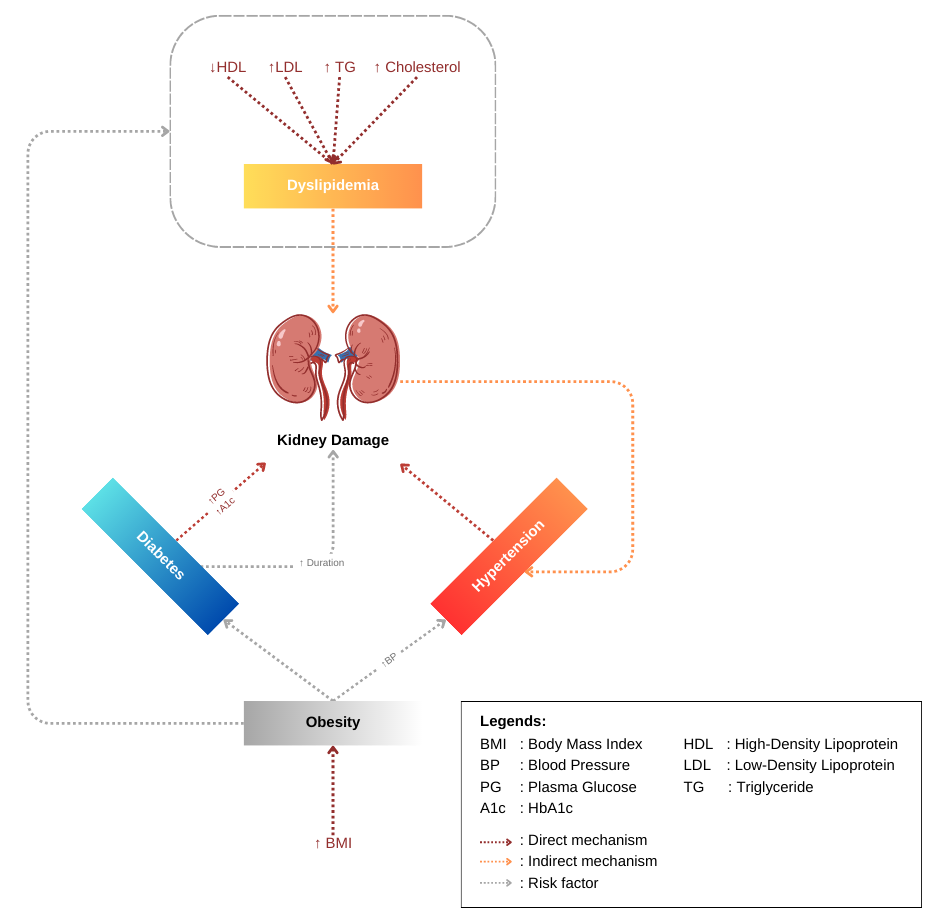}}
\caption{Simplified diabetic nephropathy mechanism}
\label{fig:diabetic_nephropathy_mechanism}
\end{figure}

The use of the aforementioned features was rationalized based on the complex nature of their interaction with kidney damage, as shown in Figure \ref{fig:diabetic_nephropathy_mechanism} \cite{agrawal2021,giunti2006,ameer2022,middleton2010}.

Figure \ref{fig:diabetic_nephropathy_mechanism} illustrates the simplified mechanism of diabetic nephropathy, with obesity playing a central role. Elevated BMI increases the likelihood of obesity, which subsequently acts as a risk factor for developing diabetes and hypertension through a complex pathway. The intricate sequence involves the increase of plasma glucose and HbA1c in diabetes, leading to microvascular damage and subsequent kidney damage. The duration of diabetes increases the risk of such damage. On the other hand, chronic high blood pressure resulting from obesity can lead to hypertension, causing microvascular damage and putting the individual at risk of kidney damage. Additionally, kidney damage can, in turn, induce hypertension, creating an inner loop-like mechanism that worsens the condition. Furthermore, obesity also serves as a risk factor for lipid profile issues, such as an increase in LDL, TG, and cholesterol, and a decrease in HDL, posing a risk of dyslipidemia. Dyslipidemia, in turn, indirectly contributes to the kidney damage.

\subsection{Design of Experiment}
This study aimed to evaluate the performance of supervised learning algorithms in predicting the risk of developing albuminuria in patients with T2DM patients. We evaluated several supervised learning algorithms, including 6 machine learning algorithms and 1 deep learning algorithm. The machine learning algorithms used were Naïve Bayes, Support Vector Machine (SVM), decision tree, random forest, AdaBoost, and XGBoost. Among these machine learning algorithms, random forest, AdaBoost, and XGBoost are ensemble algorithms. The deep learning algorithm employed in this study is the Multi-Layer Perceptron (MLP). The use of deep learning in this experimental design, as compared to machine learning algorithms, was intended to evaluate its potential performance considering the limited size of the dataset. Table \ref{tab:design_of_experiment} presents the complete experimental design used in this study. We used the scikit-learn library \cite{scikit}, version 1.0.2, as our primary machine learning and deep learning toolkit. Additionally, we employed the xgboost library \cite{xgboost}, version 1.6.2, specifically designed for implementing the XGBoost algorithm.

\begin{table}[htbp]
\caption{Design of Experiment}
\begin{center}
\begin{tabular}{|l|l|l|}
\hline
\textbf{Algorithm} & \multicolumn{2}{|l|}{\textbf{Observable Factors}} \\ \hline
\multirow{2}{*}{\text{Naïve Bayes}} & \textbf{Class} & \text{sklearn.naive\_bayes.GaussianNB} \\ 
\cline{2-3}
 & \textbf{Params} & \text{priors=None, var\_smoothing=1e-09} \\ \hline
\multirow{2}{*}{\text{SVM}} & \textbf{Class} & \text{sklearn.svm.SVC} \\ 
\cline{2-3}
 & \textbf{Params} & \text{C=1.0, kernel='rbf', max\_iter=-1} \\ \hline
\multirow{2}{*}{\text{Decision Tree}} & \textbf{Class} & \text{sklearn.tree.DecisionTreeClassifier} \\ 
\cline{2-3}
 & \textbf{Params} & \text{criterion='gini', max\_depth=None} \\ \hline
\multirow{3}{*}{\text{Random Forest}} & \textbf{Class} & \text{sklearn.ensemble.RandomForestClassifier} \\ 
\cline{2-3}
 & \textbf{Params} & \text{n\_estimators=100, criterion='gini',} \\
 & & \text{max\_depth=None} \\ \hline
\multirow{3}{*}{\text{AdaBoost}} & \textbf{Class} & \text{sklearn.ensemble.AdaBoostClassifier} \\ 
\cline{2-3}
 & \textbf{Params} & \text{n\_estimators=50, learning\_rate=1.0,} \\
 & & \text{algorithm='SAMME.R'} \\ \hline
\multirow{3}{*}{\text{XGBoost}} & \textbf{Class} & \text{xgboost.XGBClassifier} \\ 
\cline{2-3}
 & \textbf{Params} & \text{n\_estimators=2, max\_depth=1,} \\
 & & \text{learning\_rate=1} \\ \hline
\multirow{3}{*}{\text{MLP}} & \textbf{Class} & \text{sklearn.neural\_network.MLPClassifier} \\ 
\cline{2-3}
 & \textbf{Params} & \text{hidden\_layer\_sizes=(100,),} \\ 
 & & \text{learning\_rate\_init=3e-3,} \\
 & & \text{max\_iter=200} \\ \hline
\end{tabular}
\label{tab:design_of_experiment}
\end{center}
\end{table}

The dataset is split into training and test datasets using the \texttt{train\_test\_split} function provided by scikit-learn. Since the dataset size is relatively small, we opted for a train-test ratio of 0.75:0.25.

\subsection{Evaluation Strategy}
We used precision (\ref{eq:precision}), recall (\ref{eq:recall}), accuracy (\ref{eq:accuracy}), and f1-score (\ref{eq:f1_score}) as the evaluation metrics for our study. A competent model is expected to exhibit high values for precision, recall, accuracy, and f1-score.

\begin{equation}
    precision = \frac{TP}{TP + FP}
    \label{eq:precision}
\end{equation}

\begin{equation}
    recall = \frac{TP}{TP + FN}
    \label{eq:recall}
\end{equation}

\begin{equation}
    accuracy = \frac{TP + TN}{TP + FP + FN + TN}
    \label{eq:accuracy}
\end{equation}

\begin{equation}
    F1\textit{-}score = \frac{2 \times precision \times recall}{precision + recall}
    \label{eq:f1_score}
\end{equation}

\subsection{Ethics Approval}
This study has been ethically approved by the Health Ethics Committee of Cipto Mangunkusumo Hospital, Faculty of Medicine Universitas Indonesia, Jakarta, Indonesia number KET-246/UN2.F1/ETIK/PPM.00.02/2022.

\section{Result}

\begin{figure*}[htbp]
\centerline{\includegraphics[width=1\textwidth]{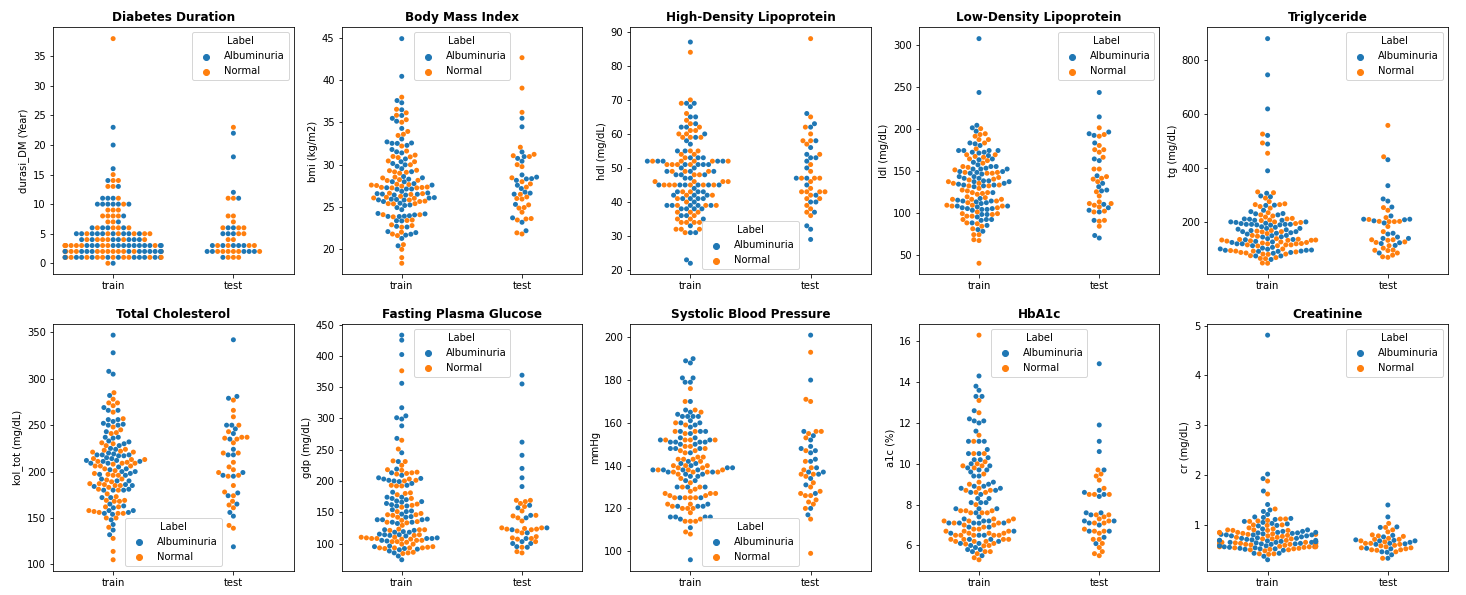}}
\caption{Distribution of the train-test dataset}
\label{fig:train_test_dataset}
\end{figure*}

\begin{figure*}[htbp]
\centerline{\includegraphics[width=0.75\textwidth]{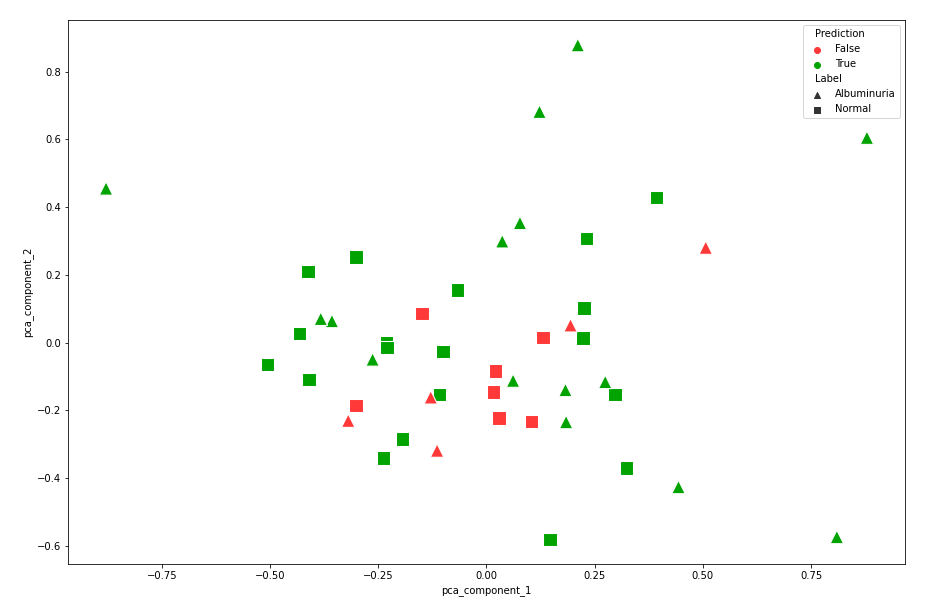}}
\caption{Error analysis}
\label{fig:error_analysis}
\end{figure*}

Figure \ref{fig:train_test_dataset} shows the distribution of the train and test datasets used in our study. The test ratio is set at 0.25, resulting in 138 records for the train dataset and 46 records for the test dataset. As depicted in Figure \ref{fig:train_test_dataset}, the dataset exhibits a relatively uniform distribution. Although several data points appear to be potential outliers, we made the decision to retain them deliberately to introduce bias to the learning algorithms and thus reduce the risk of overfitting.

We train each supervised learning algorithm with its defined parameters, as shown in Table \ref{tab:design_of_experiment}, using the training dataset, which results in trained models. These models are then tested using the test dataset, producing predicted labels. The performance of the models is evaluated by comparing these predicted labels with their corresponding true labels, which serve as the ground truth.

The model evaluation results are shown in Table \ref{tab:classification_report}. As depicted in the table, the machine learning algorithms did not yield satisfactory results. Among the various machine learning algorithms experimented with, Naïve Bayes outperformed the others. It demonstrated an accuracy of up to 0.65 when predicting the test dataset. In contrast, the remaining machine learning algorithms exhibited prediction accuracies of no more than 0.5. Even the ensemble models failed to surpass Naïve Bayes in predicting the test dataset. However, it's worth noting that none of these machine learning algorithms achieved an f1-score higher than 0.5, suggesting that the model is insufficient for predicting the risk of albuminuria in T2DM patients.

\begin{table}[htbp]
\caption{Classification Report}
\begin{center}
\begin{tabular}{|c|c|c|c|c|}
\hline
\textbf{Algorithm} & \textbf{Precision} & \textbf{Recall} & \textbf{Accuracy} & \textbf{F1-Score} \\ \hline
\text{Naïve Bayes} & \text{0.67} & \text{0.4} & \text{0.65} & \text{0.50} \\ \hline
\text{SVM} & \text{0.50} & \text{0.70} & \text{0.57} & \text{0.58} \\ \hline
\text{Decision Tree} & \text{0.45} & \text{0.50} & \text{0.52} & \text{0.48} \\ \hline
\text{Random Forest} & \text{0.50} & \text{0.60} & \text{0.57} & \text{0.55} \\ \hline
\text{AdaBoost} & \text{0.44} & \text{0.40} & \text{0.52} & \text{0.42} \\ \hline
\text{XGBoost} & \text{0.46} & \text{0.55} & \text{0.52} & \text{0.50} \\ \hline
\textbf{MLP} & \textbf{0.68} & \textbf{0.75} & \textbf{0.74} & \textbf{0.71} \\ \hline
\end{tabular}
\label{tab:classification_report}
\end{center}
\end{table}

The superior results were obtained from the deep learning algorithm, specifically the Multi-Layer Perceptron (MLP), which achieved an accuracy and f1-score of 0.74 and 0.71, respectively. This algorithm outperformed the machine learning algorithms, which only achieved accuracy and f1-scores of up to 0.65 and 0.55, respectively. Additionally, the MLP algorithm exhibited the highest precision and recall scores compared to the other algorithms, scoring 0.68 and 0.75, respectively. This result outperformed the study by Khitan et al. \cite{khitan2021} and Lin et al. \cite{lin2022}, indicating that the algorithm might be acceptable for predicting the risk of albuminuria among T2DM patients. However, further improvements are needed, particularly in terms of the dataset size and variety, to achieve better results.

To gain a better understanding of the model evaluation results, we conducted a visual error analysis on the prediction outcomes of the MLP model. To facilitate visualization, we used the Principal Component Analysis (PCA) method to reduce the features from 10 to 2 dimensions. Subsequently, we used square and triangle markers to represent the normal and albuminuria labels, respectively, while using red and green colors to indicate false and true predictions, respectively. The visualization of the model evaluation results can be observed in Figure \ref{fig:error_analysis}.

As shown in Figure \ref{fig:error_analysis}, the false predictions could have either a normal or albuminuria label. However, the interesting point revealed by the visualization in the figure is that the falsely predicted labels are spread out but relatively close to the adjacent cluster that forms the true predictions. This indicates that the data characteristics of 'normal' and 'albuminuria' at some points have little difference, which might cause the algorithm to experience difficulty in creating separate boundaries between the labels, leading to false predictions and resulting in lower accuracy. This phenomenon may be explained by the nature of the patient data.

For several patients, the risk of developing albuminuria might not be strongly correlated with the features in the dataset. For example, there could be a patient with uncontrolled diabetes, indicated by high blood glucose and high lipid profile, but not developing albuminuria, while another patient with normal glucose levels and normal lipid profile developing albuminuria. This complexity arises because the human body is complex, and the risk of developing a disease may be influenced by multiple risk factors that are not apparent in the dataset. Therefore, one possible solution to improve the model's accuracy is to increase the dataset size and variety, allowing the learning algorithms to better understand the complex patterns present in such data.

Despite the complex nature of the patient data, the outperforming performance of the MLP algorithm might be beneficial due to its architecture. The MLP might consist of several to even thousands of hidden layers. The MLP uses the backpropagation algorithm to update its weights based on the learned data \cite{mitchell1997}. This enables the MLP to solve complex problems relatively easily compared to other traditional machine learning algorithms when dealing with such complex data.

\section{Conclusion}
We have developed a supervised learning model to predict the risk of developing albuminuria in patients with T2DM. Among the various supervised learning models examined in this study, the MLP algorithm demonstrated superior performance in terms of precision, recall, accuracy, and f1-score. Specifically, the algorithm achieved values of 0.68, 0.75, 0.74, and 0.71 for precision, recall, accuracy, and f1-score, respectively. To further enhance the model's performance, we recommend augmenting the dataset with additional data to increase its size and diversity. Additionally, we propose conducting further research into the utilization of deep learning algorithms like MLP to effectively handle the complexities inherent in patient data.

\section*{Acknowledgment}
This study was conducted as part of the ID-CALDERA (Indonesia's Cardiovascular Risk
Stratification in Diabetes Using Smartphone-Based Retinal
Imaging) project, which aims to develop AI-based models and technologies for predicting diabetic complications at an earlier stage. The vision and mission of ID-CALDERA are centered on reducing the impact and burden of diabetic complications. We extend our gratitude to everyone who has provided support for our work, whether individually or institutionally.


\begin{thebibliography}{00}
\bibitem{perkeni2021} Pengurus Besar Perkumpulan Endokrinologi Indonesia, \textit{Pedoman Pencegahan dan Pengelolaan Diabetes Mellitus Tipe 2 Dewasa di Indonesia 2021}. Jakarta, Indonesia: Pengurus Besar Perkumpulan Endokrinologi Indonesia, 2021.

\bibitem{idf2021} International Diabetes Federation, \textit{IDF Atlas}. Brussels, Belgium: International Diabetes Federation, 2021.

\bibitem{kyoru2020} I. Kyrou, C. Tsigos, C. Mavrogianni, et al., "Sociodemographic and Lifestyle-related Risk Factors for Identifying Vulnerable Groups for Type 2 Diabetes: A Narrative Review with Emphasis on Data from Europe," in \textit{BMC Endocrine Disorders}, vol. 20, no. Suppl 1, p. 134, 2020. [Online]. doi: 10.1186/s12902-019-0463-3.

\bibitem{chamine2022} I. Chamine, J. Hwang, S. Valenzuela, M. Marino, A. E. Larson, J. Georgescu, M. Latkovic-Taber, H. Angier, J. E. DeVoe, and N. Huguet, "Acute and Chronic Diabetes-Related Complications Among Patients With Diabetes Receiving Care in Community Health Centers," in \textit{Diabetes Care}, vol. 45, no. 10, pp. e141-e143, 2022. doi: 10.2337/dc22-0420.

\bibitem{haas1993} L. B. Haas, "Chronic Complications of Diabetes Mellitus," in \textit{Nursing Clinics of North America}, vol. 28, no. 1, pp. 71-85, 1993, doi: 10.1016/S0029-6465(22)02837-7.

\bibitem{kdigo2012} KDIGO. "2012 clinical practice guideline for the evaluation and management of chronic kidney disease," in \textit{Kidney International Supplements}, vol. 3, pp. 1-150, 2013.

\bibitem{pavkov2018} M. E. Pavkov, A. J. Collins, J. Coresh, et al., "Kidney Disease in Diabetes," in \textit{Diabetes in America}, 3rd edition, C. C. Cowie, S. S. Casagrande, A. Menke, et al., Eds., USA: National Institute of Diabetes and Digestive and Kidney Diseases, August 2018. [Online]. Available: https://www.ncbi.nlm.nih.gov/books/NBK568002/. Accessed: July 24, 2023

\bibitem{khitan2021} Z. Khitan, T. Nath, P Santhanam, "Machine Learning Approach to Predicting Albuminuria in Persons with Type 2 Diabetes: An Analysis of the LOOK AHEAD Cohort," in \textit{The Journal of Clinical Hypertension}, vol. 23, no. 12, pp. 2137-2145, 2021, doi: 10.1111/jch.14397.

\bibitem{lin2022} W. Lin, et al., "Development of a Risk Model for Predicting Microalbuminuria in the Chinese Population Using Machine Learning Algorithms," in \textit{Frontiers in Medicine}, vol. 9, 2022, doi: 10.3389/fmed.2022.775275

\bibitem{agrawal2021} R. Agarwal, "Pathogenesis of Diabetic Nephropathy," in \textit{ADA Clinical Compendia}, vol. 2021, no. 1, pp. 2–7, June 1, 2021. doi: 10.2337/db20211-2

\bibitem{giunti2006} S. Giunti, D. Barit, and M. E. Cooper, "Mechanisms of Diabetic Nephropathy," in \textit{Hypertension}, vol. 48, no. 4, pp. 519-526, 2006, doi: 10.1161/01.HYP.0000240331.32352.0c. 

\bibitem{ameer2022} O.Z. Ameer, "Hypertension in Chronic Kidney Disease: What Lies Behind the Scene," \textit{Frontiers in Pharmacology}, vol. 13, p. 949260, Oct. 11, 2022, doi: 10.3389/fphar.2022.949260.

\bibitem{middleton2010} J. P. Middleton and P. H. Pun, "Hypertension, Chronic Kidney Disease, and the Development of Cardiovascular Risk: A Joint Primacy," in \textit{Kidney International}, vol. 77, no. 9, pp. 753-755, 2010. doi: 10.1038/ki.2010.19.

\bibitem{scikit} F. Pedregosa, G. Varoquaux, A. Gramfort, V. Michel, B. Thirion, O. Grisel, M. Blondel, P. Prettenhofer, R. Weiss, V. Dubourg, J. Vanderplas, A. Passos, D. Cournapeau, M. Brucher, M. Perrot, and É. Duchesnay, "Scikit-learn: Machine Learning in Python," in \textit{Journal of Machine Learning Research}, vol. 12, no. 85, pp. 2825-2830, 2011.

\bibitem{xgboost} T. Chen and C. Guestrin, "XGBoost: A Scalable Tree Boosting System," in \textit{Proceedings of the 22nd ACM SIGKDD International Conference on Knowledge Discovery and Data Mining (KDD '16)}, San Francisco, California, USA, 2016, pp. 785-794, doi: 10.1145/2939672.2939785.

\bibitem{mitchell1997} T. Mitchell, \textit{Machine Learning}. USA: McGraw-Hill, 1997.
\end{thebibliography}
\end{document}